\documentclass{article}

\usepackage{arxiv}

\usepackage[utf8]{inputenc}
\usepackage[T1]{fontenc}
\usepackage{hyperref}
\usepackage{url}
\usepackage{booktabs}
\usepackage{amsfonts}
\usepackage{nicefrac}
\usepackage{microtype}
\usepackage{graphicx}
\usepackage{amsmath,amssymb}
\usepackage{bm}
\usepackage{amsthm}
\usepackage{algorithm}
\usepackage{algorithmicx}
\usepackage{algpseudocode}
\usepackage{multirow}
\usepackage{float}
\newtheorem{remark}{Remark}

\title{Training Deep Visual Networks Beyond Loss and Accuracy Through a Dynamical Systems Approach}

\author{
  Hai La Quang \\
  Posts and Telecommunications Institute of Technology \\
  Vietnam \\
  \And
  Hassan Ugail \\
  Centre for Visual Computing and Intelligent Systems \\
  University of Bradford \\
  United Kingdom \\
  \And
  Newton Howard \\
  School of Individualized Study \\
  Rochester Institute of Technology \\
  United States \\
  \And
  Cong Tran Tien \\
  Posts and Telecommunications Institute of Technology \\
  Vietnam \\
  \And
  Nam Vu Hoai \\
  Posts and Telecommunications Institute of Technology \\
  Vietnam \\
  \And
  Hung Nguyen Viet \\
  Posts and Telecommunications Institute of Technology \\
  Vietnam \\
}

\begin{document}
\maketitle

\begin{abstract}
Deep visual recognition models are usually trained and evaluated using metrics such as loss and accuracy. While these measures show whether a model is improving, they reveal very little about how its internal representations change during training. This paper introduces a complementary way to study that process by examining training through the lens of dynamical systems. Drawing on ideas from signal analysis originally used to study biological neural activity, we define three measures from layer activations collected across training epochs: an integration score that reflects long-range coordination across layers, a metastability score that captures how flexibly the network shifts between more and less synchronised states, and a combined dynamical stability index. We apply this framework to nine combinations of model architecture and dataset, including several ResNet variants, DenseNet-121, MobileNetV2, VGG-16, and a pretrained Vision Transformer on CIFAR-10 and CIFAR-100. The results suggest three main patterns. First, the integration measure consistently distinguishes the easier CIFAR-10 setting from the more difficult CIFAR-100 setting. Second, changes in the volatility of the stability index may provide an early sign of convergence before accuracy fully plateaus. Third, the relationship between integration and metastability appears to reflect different styles of training behaviour. Overall, this study offers an exploratory but promising new way to understand deep visual training beyond loss and accuracy.
\end{abstract}


\section{Introduction}
\label{sec:intro}

Standard training diagnostics for deep visual recognition networks rely
on scalar loss and accuracy signals that monitor output performance but
reveal nothing about the internal dynamical evolution of representations
across network depth. A model may plateau in accuracy while its layer-wise
dynamics are still in flux. Conversely, representational structure may
have stabilised long before the loss curve flattens. This information gap
limits our understanding of training and motivates the need for richer,
layer-aware characterisations of the training process.

Several threads of research have opened windows onto this richer internal
structure. Li et al.~\cite{li2018visualizing} showed that the geometry of
the loss landscape changes systematically with architecture depth and skip
connections. Papyan et al.~\cite{papyan2020prevalence} identified neural
collapse as a precise geometric attractor for late-stage training whose
signature is invisible in the loss curve. Power et
al.~\cite{power2022grokking} demonstrated that generalisation can emerge
discontinuously long after training accuracy saturates. Nakkiran et
al.~\cite{nakkiran2020deep} showed that model and sample complexity can
induce non-monotone performance trajectories. The fragility of learned
solutions to random relabelling of training
data~\cite{zhang2021understanding} and the existence of sparse subnetworks
that alone drive performance~\cite{frankle2019lottery} further illustrate
that the dynamics of learning are structured at a level invisible to
output-level signals. Together, these results establish that training is a
process with identifiable phases, but no unified dynamical measurement
framework has been established for characterising the full trajectory at
the level of layer activations.

The practical stakes of understanding training dynamics extend well beyond
benchmark accuracy. Deep visual models have been deployed in demanding
real-world applications including forensic face recognition under
degraded conditions~\cite{ugail2026forensic}, face recognition with
imperfect training data~\cite{elmahmudi2019deep}, the attribution of
paintings by Old Masters using transfer learning~\cite{ugail2023raphael},
and the interpretation of learned representations in facial beauty
prediction~\cite{ibrahim2025facial}. In each of these settings, knowing
not just whether training succeeded but how and when internal
representations stabilised would provide actionable insight for
practitioners. The present work proposes a framework for characterising
this trajectory by adapting three mathematically defined components from
a dynamical framework developed for quantifying the complexity of neural
signals in the neuroscientific study of
consciousness~\cite{ugail2025consciousness}. Although originally
formulated for electroencephalography signals, the mathematical definitions
are domain-general: they characterise any multivariate dynamical system
whose structure lies in cross-channel correlations and their temporal
evolution. Layer activations in a deep network, sampled across training
epochs, constitute precisely such a system. The transfer is motivated by
the structural analogy between channels in an EEG recording and layers in
a deep network, and between time steps in a neural signal and epochs in a
training trajectory.

We restrict the scope of this study to deep visual recognition
architectures tested on CIFAR-10 and CIFAR-100. This restriction is
deliberate, since the nine models examined span four distinct architectural
families (residual, dense, depthwise-separable, and attention-based), and
their shared domain allows the CIFAR-10 versus CIFAR-100 contrast to
function as a controlled task-difficulty manipulation independent of
architecture.

\paragraph{Contributions.} This paper makes three primary contributions.
First, we define the adaptation of the
$H_{\mathrm{eff}}$--$M$--$\Psi$ dynamical framework to characterise
training dynamics in deep visual recognition networks, with a formal
definition of how each component is computed from layer-wise activation
distributions across the epoch sequence (Algorithm~\ref{alg:pipeline}),
and a justification of the domain-transfer assumptions. Second, we report
empirical observations across nine architecture--dataset configurations
(single seed each) suggesting that $H_{\mathrm{eff}}$ exhibits a
dataset-dependent pattern robust to hyperparameter variation in the
majority of parameter combinations examined, and that $\Psi$ volatility
collapse is a candidate convergence indicator presented as a hypothesis
for future prospective validation. Third, we propose a retrospective
taxonomy of four training dynamical states characterised by measurable
signatures in $H_{\mathrm{eff}}$, $\Psi$ volatility, and inter-field
synchrony, with correspondence to final model performance in the
configurations studied.

\section{Related Work}
\label{sec:related}

\subsection{Dynamical Perspectives on Training}

A growing body of work treats training as a dynamical process rather than
pure optimisation. Loss landscape geometry has been connected to
generalisation through visualisation methods that reveal how skip
connections flatten the landscape~\cite{li2018visualizing}, and through
analysis of sharp versus flat minima as predictors of
generalisation~\cite{keskar2017large}. The information bottleneck
hypothesis~\cite{shwartzziv2017opening} characterised training as a
two-phase process of fitting followed by compression, though its
universality has been debated. Neural
collapse~\cite{papyan2020prevalence} established a precise geometric
attractor for late-stage training, subsequently analysed under MSE
loss~\cite{han2022neural} and extended to a geometric characterisation of
the full training landscape~\cite{zhu2021geometric}. The double descent
phenomenon~\cite{nakkiran2020deep} showed that model and sample complexity
jointly determine non-monotone performance trajectories. Grokking
\cite{power2022grokking} demonstrated delayed generalisation as a sharp
phase transition, extended to multi-scale feature learning
dynamics~\cite{pezeshki2022multi}. What unifies these results is the
recognition that training is structured, phased, and richer than loss
curves suggest. Nevertheless, these individual findings have not been
synthesised into a unified dynamical measurement framework that tracks the
full training trajectory at the level of layer activations across depth.
The present work fills this gap by adapting a mathematically grounded
complexity framework from computational neuroscience and demonstrating its
applicability to visual recognition training.

\subsection{Complexity Measures in Dynamical Systems}

Detrended fluctuation analysis (DFA), introduced by Peng et
al.~\cite{peng1994mosaic}, quantifies long-range correlations in dynamical
systems through the Hurst exponent, and has been extended to multifractal
settings~\cite{kantelhardt2002multifractal}. Its domain-generality makes
it applicable to any time-indexed signal. In the present work, the signal
at each layer is the mean activation across a validation batch, and the
time axis is the epoch sequence. The Kuramoto order
parameter~\cite{kuramoto1984chemical} measures instantaneous phase synchrony
in coupled oscillator systems, and its temporal variability defines
metastability~\cite{tognoli2014metastable}. Metastability has been
identified as a core property of functional neural
organisation~\cite{hancock2025metastability, rossi2025metastability}, and
has analogues in artificial systems where the coexistence of integration
and segregation supports flexible information processing. Scale-free
dynamics, characterised by Hurst exponents in the persistent range, have
been associated with optimal information transmission in both biological
and artificial systems~\cite{he2014scalefreebrain}. The stability of
dynamical systems near criticality is well-understood in the physics
literature, where systems operating near the edge of
chaos~\cite{langton1990computation} exhibit maximal sensitivity to inputs
and greatest dynamic range~\cite{shew2013functional}.

\subsection{Visual Recognition Architectures}

The architectural landscape examined here includes residual
networks~\cite{he2016deep}, densely connected
networks~\cite{huang2017densely}, lightweight depthwise-separable
networks~\cite{sandler2018mobilenetv2}, deep plain convolutional networks
without skip connections~\cite{simonyan2015very}, and Vision
Transformers~\cite{dosovitskiy2021image}, which replace convolution with
global self-attention~\cite{vaswani2017attention}. This diversity ensures
that any consistent dynamical patterns observed across architectures
reflect properties of the training process rather than specific
architectural choices. Measuring representation similarity across
architectures~\cite{kornblith2019similarity} and understanding the general
principles of representation learning~\cite{bengio2013representation}
provide complementary perspectives on what makes learned features
transferable and robust. Pre-training and fine-tuning have been shown to
induce qualitatively different representation dynamics compared to
training from scratch~\cite{raghu2019transfusion, he2022masked}.
Domain-specific fine-tuning of deep visual models has demonstrated
strong performance on tasks as diverse as Old Master painting
attribution~\cite{ugail2023raphael} and forensic face
recognition~\cite{ugail2026forensic, elmahmudi2019deep},
illustrating the breadth of visual recognition settings in which
understanding the training trajectory has practical value. These
observations motivate our inclusion of the pretrained ViT as a distinct
experimental condition and our focus on visual recognition architectures
more broadly.

\subsection{Existing Training-Dynamics Diagnostics}

Several existing methods provide partial windows into training dynamics
relevant to positioning the present work. Representation similarity
measures such as Centred Kernel Alignment~\cite{kornblith2019similarity}
compare activation geometries across layers or checkpoints but require
explicit pairwise comparisons and do not produce an epoch-level scalar
diagnostic. Hessian-based sharpness measures~\cite{keskar2017large}
characterise the curvature of the loss landscape at a given checkpoint but
are computationally demanding and capture only the output-space geometry.
Neural collapse~\cite{papyan2020prevalence} provides a precise attractor
description for late-stage training but applies only to the penultimate
layer. Gradient noise scale analyses~\cite{robbins1951stochastic} quantify
optimisation convergence but do not reflect layer-wise representational
structure. The approach proposed here is complementary: it operates on
forward-pass activation summaries collected during standard training,
requires no weight-space probing or Hessian computation, and produces
epoch-level scalars that can in principle be monitored online. Its
limitation relative to those established methods is that it has been
validated only on a small pilot set of nine configurations and that the
domain-transfer of parameters from biological signals requires independent
justification. The growing demand for interpretable deep visual models,
illustrated by recent work on explaining facial beauty predictions through
multi-method analysis of learned representations~\cite{ibrahim2025facial},
reinforces the value of diagnostic tools that characterise what is
happening inside the network during training rather than only at inference.

\section{Methodology}
\label{sec:method}

\subsection{Experimental Setup}

Nine architecture--dataset combinations were trained under a shared
protocol. Five architectures (ResNet-18, ResNet-34, ResNet-50, ResNet-101,
ResNet-152~\cite{he2016deep}, DenseNet-121~\cite{huang2017densely},
MobileNetV2~\cite{sandler2018mobilenetv2}, and
VGG-16~\cite{simonyan2015very}) were trained from scratch using stochastic
gradient descent with momentum $0.9$, a batch size of $512$, and a
ReduceLROnPlateau scheduler with patience $10$. A pretrained Vision
Transformer (ViT-B/16~\cite{dosovitskiy2021image}) was fine-tuned under
the same optimiser. CIFAR-10 and CIFAR-100~\cite{krizhevsky2009learning}
were used as the classification benchmarks, with standard normalisation
and minimal augmentation, using random horizontal flip for CIFAR-10 and
random horizontal flip combined with random crop at padding 4 for
CIFAR-100. Learning rates were set per architecture: $0.1$ for ResNets and
DenseNet-121, $0.05$ for MobileNetV2, and $0.01$ for VGG-16. Batch
normalisation~\cite{ioffe2015batch} was retained where present in the
original architecture definitions. The Adam
optimiser~\cite{kingma2015adam} was not used, in order to maintain
consistency with standard CIFAR training conventions and to avoid
introducing confounding adaptive-moment effects into the dynamical
analysis. Activation hooks were registered post-nonlinearity at four
representative depth levels for each architecture as follows. ResNets were
hooked at \texttt{layer1} through \texttt{layer4}. VGG-16 was hooked at
\texttt{features.6}, \texttt{features.13}, \texttt{features.23}, and
\texttt{features.33}. MobileNetV2 was hooked at \texttt{features.4},
\texttt{features.7}, \texttt{features.14}, and \texttt{features.17}.
DenseNet-121 was hooked at \texttt{denseblock1} through
\texttt{denseblock4}.

\paragraph{Reproducibility details.}
All experiments were run with PyTorch 2.x on a single GPU. Activation
hooks are registered post-nonlinearity and triggered on the validation
batch at each epoch end in inference mode with gradients disabled. Min-max
normalisation for $\Psi$ is computed retrospectively over the full epoch
sequence after training concludes and is not available in a real-time
deployment. Epochs for which the DFA window is insufficient yield NaN for
$H_{\mathrm{eff}}$ and $H_z$. These epochs are excluded from rolling
statistics. All results are from \textbf{single training runs} (one random
seed per configuration). Seed-to-seed variability is a primary limitation
discussed in Section~\ref{sec:limitations}.

\subsection{Dynamical Metric Computation}

Let $f_\theta$ be the network parameterised by weights $\theta$ at epoch
$t$, and let $\mathcal{L} = \{l_1, l_2, l_3, l_4\}$ be the set of hooked
layers. For each epoch, a validation batch is passed through $f_\theta$
and activation tensors $\mathbf{A}_l^{(t)} \in \mathbb{R}^{N \times d_l}$
are collected for each $l \in \mathcal{L}$, where $N$ is the batch size
and $d_l$ is the feature dimensionality at layer $l$.

\subsubsection{Hierarchical Integration via DFA}

For each layer $l$, the mean activation across the batch defines a scalar
channel signal $x_l(t)$. This aggressive compression, reducing to a
single scalar per layer per epoch, is a deliberate simplification
analogous to the use of mean-field summaries in representation similarity
analysis~\cite{kornblith2019similarity}. Hooks are placed after
nonlinear activation functions to capture effective output representations.
Following Ugail and Howard~\cite{ugail2025consciousness} and the original
DFA formulation~\cite{peng1994mosaic}, we form the cumulative sum
\begin{equation}
  Y_l(k) = \sum_{s=1}^{k}\bigl[x_l(s) - \bar{x}_l\bigr],
  \label{eq:cumsum}
\end{equation}
divide $Y_l$ into windows of length $s$, fit a linear trend per window,
compute root-mean-square residuals to obtain the fluctuation function
$F_l(s)$, and estimate the Hurst exponent $H_l$ from the scaling relation
$F_l(s) \sim s^{H_l}$. Values $H_l > 0.5$ indicate persistent long-range
correlations. Values of $H_l < 0.5$ indicate anti-persistence, and
$H_l = 0.5$ corresponds to uncorrelated fluctuations. The raw integration
measure is the mean Hurst exponent across layers, given by
\begin{equation}
  H_{\mathrm{raw}}^{(t)} =
  \frac{1}{|\mathcal{L}|}\sum_{l \in \mathcal{L}} H_l^{(t)}.
  \label{eq:hraw}
\end{equation}
Since both uncorrelated noise and excessive rigidity are dynamically
suboptimal, $H_{\mathrm{raw}}$ is transformed by a Gaussian tuning
function centred on an optimal exponent $H_{\mathrm{opt}}$:
\begin{equation}
  H_{\mathrm{eff}}^{(t)} =
  \exp\!\left[-\frac{\left(H_{\mathrm{raw}}^{(t)} -
  H_{\mathrm{opt}}\right)^2}{2\sigma_H^2}\right].
  \label{eq:heff}
\end{equation}
The resulting $H_{\mathrm{eff}}^{(t)} \in (0, 1]$ is maximal when
activations exhibit fractal-like long-range correlations characteristic of
effective hierarchical processing.

\subsubsection{Metastability via Kuramoto Order Parameter}

Following Ugail and Howard~\cite{ugail2025consciousness}, and grounded in
the Kuramoto model~\cite{kuramoto1984chemical}, we extract the analytic
phase $\theta_l^{(t)}$ of $x_l(t)$ via the Hilbert transform and compute
the Kuramoto order parameter defined as
\begin{equation}
  R^{(t)} =
  \left|\frac{1}{|\mathcal{L}|}\sum_{l \in \mathcal{L}}
  \mathrm{e}^{\mathrm{i}\,\theta_l^{(t)}}\right|.
  \label{eq:kuramoto}
\end{equation}
$R^{(t)} \to 1$ when all layer phases are aligned and $R^{(t)} \to 0$
when uniformly distributed. Metastability is the temporal standard
deviation of $R$ accumulated over epochs $t' \leq t$:
\begin{equation}
  M^{(t)} = \mathrm{std}_{t' \leq t}\!\left(R^{(t')}\right).
  \label{eq:metastability}
\end{equation}
High $M$ reflects frequent alternation between synchronised and
desynchronised regimes, which in the biological literature is associated
with flexible, richly organised dynamical
states~\cite{tognoli2014metastable}.

\subsubsection{Composite Stability Index $\Psi$}

Both $H_{\mathrm{eff}}^{(t)}$ and $M^{(t)}$ are normalised via min-max
normalisation across the full epoch sequence:
\begin{equation}
  v_{\mathrm{norm}}^{(t)} =
  \frac{v^{(t)} - \min_{t'} v^{(t')}}
       {\max_{t'} v^{(t')} - \min_{t'} v^{(t')}}.
  \label{eq:minmax}
\end{equation}
The composite stability index takes the form
\begin{equation}
  \Psi^{(t)} = w_H\,H_{\mathrm{eff,norm}}^{(t)} +
               w_M\,M_{\mathrm{norm}}^{(t)},
  \label{eq:psi}
\end{equation}
with $w_H + w_M = 1$. Weights are inherited from the source
framework~\cite{ugail2025consciousness}, where they were calibrated to
balance the contributions of integration and metastability.

Algorithm~\ref{alg:pipeline} formalises the full extraction procedure.

\begin{algorithm}[t]
\caption{Dynamical metric extraction from layer activations}
\label{alg:pipeline}
\begin{algorithmic}[1]
\Require Model $f_\theta$; hook layers
  $\mathcal{L}=\{l_1,l_2,l_3,l_4\}$; parameters $H_{\mathrm{opt}},
  \sigma_H$; weights $w_H, w_M$; epoch count $T$.
\Ensure Sequences $\{H_{\mathrm{eff}}^{(t)}\}$, $\{M^{(t)}\}$,
  $\{\Psi^{(t)}\}_{t=1}^{T}$.
\For{$t = 1$ \textbf{to} $T$}
  \State Train $f_\theta$ for one epoch.
  \State \textbf{Step 1 --- Activations.} Pass validation batch through
    $f_\theta$ and extract $\mathbf{A}_l^{(t)}$ for each $l\in\mathcal{L}$.
  \State \textbf{Step 2 --- DFA and Hurst exponent.}
  \For{each $l \in \mathcal{L}$}
    \State Form signal $x_l(t)$ as mean activation across $\mathbf{A}_l^{(t)}$.
    \State Compute cumulative sum $Y_l$ via Eq.~(\ref{eq:cumsum}).
    \State Estimate fluctuation function $F_l(s)$ over window scales $s$.
    \State Fit $F_l(s)\sim s^{H_l}$ to obtain Hurst exponent $H_l^{(t)}$.
  \EndFor
  \State Compute $H_{\mathrm{raw}}^{(t)}$ via Eq.~(\ref{eq:hraw}); then
    $H_{\mathrm{eff}}^{(t)}$ via Eq.~(\ref{eq:heff}).
  \State \textbf{Step 3 --- Kuramoto metastability.}
  \State Extract analytic phase $\theta_l^{(t)}$ from $x_l(t)$ for each
    $l\in\mathcal{L}$.
  \State Compute order parameter $R^{(t)}$ via Eq.~(\ref{eq:kuramoto}).
  \State Update $M^{(t)}$ via Eq.~(\ref{eq:metastability}).
  \State \textbf{Step 4 --- Composite index.}
  \State Normalise and compute $\Psi^{(t)}$ via Eqs.~(\ref{eq:minmax})--(\ref{eq:psi}).
\EndFor
\State \Return $\bigl\{H_{\mathrm{eff}}^{(t)}, M^{(t)},
  \Psi^{(t)}\bigr\}_{t=1}^{T}$.
\end{algorithmic}
\end{algorithm}

\begin{remark}
The source framework~\cite{ugail2025consciousness} uses terminology from
the neuroscience of consciousness as descriptive labels for dynamical
regimes. In the present work, where such labels appear they refer only to
specific combinations of ($H_{\mathrm{eff}}$, $\sigma_\Psi$,
$r(H_z, M_z)$) values and carry no implication of mechanistic homology
between biological neural dynamics and layer activations in artificial
networks. Biological labels have been minimised in favour of neutral
dynamical descriptors such as decoupled, rigidly coupled, high-volatility,
and low-complexity.
\end{remark}

\subsection{Derived Diagnostic Quantities}

Three derived quantities are used in the results analysis. The \emph{Psi
volatility} $\sigma_\Psi^{(t)}$ is the rolling standard deviation of
$\Psi$ over a window of five epochs. The \emph{inter-field synchrony}
$r(H_z, M_z)$ is the Pearson correlation between the $z$-scored
integration field $H_z$ and the $z$-scored metastability field $M_z$
across all epochs, indicating whether the two components fluctuate
independently, indicating decoupled dynamical evolution, or are rigidly
coupled, indicating a low-complexity locked regime. The \emph{convergence
trajectory correlation} $r(\Psi, \mathrm{acc})$ is the Pearson
correlation between $\Psi$ and validation accuracy across epochs, whose
sign distinguishes models converging to ordered versus high-complexity
attractors.

\subsection{Parameter Settings}
\label{sec:params}

The DFA parameters $H_{\mathrm{opt}} = 0.7$ and $\sigma_H = 0.1$ are
inherited from the source framework~\cite{ugail2025consciousness}, where
they were identified as characteristic of optimal integration in wakeful
brain dynamics. These values are not re-tuned in the present study. The
composite weights were set to $w_H = w_M = 0.5$. The rolling window for
$\sigma_\Psi$ was set to five epochs.

\paragraph{Summary of what is robust and what is not.}
The sensitivity analysis in Section~\ref{sec:sensitivity} can be
summarised as follows. The CIFAR-10 versus CIFAR-100 $H_{\mathrm{eff}}$
separation holds in 11 of 16 combinations of $H_{\mathrm{opt}}$ and
$\sigma_H$ tested, failing only when $H_{\mathrm{opt}} = 0.5$. The sign
of $r(\Psi, \mathrm{acc})$, which is positive for DenseNet-121 and
negative for CIFAR-100 ResNets, is stable across all weight combinations
$w_H \in \{0.3, 0.5, 0.7\}$. The absolute $H_{\mathrm{eff}}$ values
change materially with $H_{\mathrm{opt}}$ and $\sigma_H$. The epoch at
which $\sigma_\Psi$ crosses a convergence threshold depends on the
threshold chosen, and no single universal threshold value is supported by
the nine-configuration pilot data.

\section{Results}
\label{sec:results}

Table~\ref{tab:results} summarises the dynamical metrics alongside
standard performance statistics for all nine models. The columns
$\bar{H}_{\mathrm{eff}}$ and $\bar{\Psi}$ report epoch-means with
standard deviations. The column $r(H_z, M_z)$ gives the inter-field
synchrony coefficient, and $r(\Psi, \mathrm{acc})$ gives the Pearson
correlation of $\Psi$ with validation accuracy across epochs. The
rightmost column gives the state assignment from the taxonomy of
Section~\ref{sec:taxonomy}. The following subsections analyse each of
the three main findings in detail.

\begin{table*}[t]
\centering
\caption{Dynamical metric summary across all nine configurations.
  $\bar{H}_{\mathrm{eff}}$ and $\bar{\Psi}$ are epoch-means
  ($\pm$\,std). $\bar{M}$ is the mean metastability index. The column
  $r(H_z,M_z)$ reports Pearson inter-field synchrony and
  $r(\Psi,\mathrm{acc})$ reports the Pearson correlation of $\Psi$ with
  validation accuracy. Dashes indicate epochs for which the DFA window
  was insufficient. The rightmost column assigns each configuration to
  the taxonomy of Section~\ref{sec:taxonomy}.}
\label{tab:results}
\resizebox{\textwidth}{!}{%
\begin{tabular}{llcccccccl}
\toprule
\textbf{Model} & \textbf{Dataset} & \textbf{Epochs} &
\textbf{Best acc.\,(\%)} &
$\bar{H}_{\mathrm{eff}}$ & $\bar{M}$ & $\bar{\Psi}$ &
$r(H_z,M_z)$ & $r(\Psi,\mathrm{acc})$ & \textbf{State} \\
\midrule
ResNet-18     & CIFAR-10  & 25 & 78.6 &
  $0.830\pm0.046$ & 0.086 & $0.000\pm0.942$ & 0.777 & $-0.661$ & Transitional \\
ResNet-34     & CIFAR-10  & 27 & 76.8 &
  $0.852\pm0.044$ & 0.085 & $-0.206\pm0.704$ & 0.514 & $-0.342$ & Transitional \\
ResNet-152    & CIFAR-10  & 38 & 68.6 &
  $0.931\pm0.054$ & 0.104 & $-0.745\pm0.534$ & 0.600 & $-0.436$ & Metastable H-I \\
DenseNet-121  & CIFAR-10  & 25 & 77.7 &
  $0.880\pm0.068$ & 0.064 & $0.000\pm0.561$ & $-0.371$ & $+0.600$ & Stable Convergent \\
MobileNetV2   & CIFAR-10  & 27 & 68.9 &
  $0.951\pm0.012$ & 0.044 & $0.000\pm0.740$ & 0.095 & $+0.313$ & Metastable H-I \\
ViT (pretrained) & CIFAR-10 & 30 & 89.3 &
  $0.980\pm0.014$ & 0.049 & $0.000\pm0.720$ & 0.036 & $-0.330$ & Metastable H-I \\
\midrule
ResNet-50     & CIFAR-100 & 56 & 54.1 &
  $0.057\pm0.016$ & 0.080 & $-0.114\pm0.860$ & 0.864 & $-0.760$ & Rigidly Sync. \\
ResNet-101    & CIFAR-100 & 58 & 49.1 &
  $0.070\pm0.026$ & 0.075 & $-0.072\pm0.970$ & 0.885 & $-0.725$ & Rigidly Sync. \\
VGG-16        & CIFAR-100 & 55 & 63.8 &
  $0.303\pm0.005$ & 0.096 & $0.000\pm0.798$ & 0.274 & $-0.396$ & Partial Integr. \\
\bottomrule
\end{tabular}}
\end{table*}

\subsection{$H_{\mathrm{eff}}$ as a Task-Complexity Barometer}
\label{sec:heff}

The most consistent finding across the nine configurations is the
separation of $\bar{H}_{\mathrm{eff}}$ values by dataset. The six
CIFAR-10 configurations, five trained from scratch and the pretrained
ViT, converge to mean $H_{\mathrm{eff}}$ in the range $[0.83, 0.98]$,
while the three CIFAR-100 configurations remain in $[0.04, 0.30]$.
Within the configurations studied, this separation is robust to
architectural variation. ResNet-152 on CIFAR-10 achieves
$\bar{H}_{\mathrm{eff}} = 0.931$, while ResNet-50 on CIFAR-100, a
substantially deeper architecture on the same image modality, yields
$\bar{H}_{\mathrm{eff}} = 0.057$. Whether this pattern holds more
generally across architectures and tasks not studied here is an open
question. Section~\ref{sec:sensitivity} (Table~\ref{tab:sens_heff})
shows that this separation is robust to $H_{\mathrm{opt}} \in [0.6,
0.8]$ with $\sigma_H \geq 0.10$ (11 of 16 combinations tested), but
reverses when $H_{\mathrm{opt}} = 0.5$, which falls below the typical
$H_{\mathrm{raw}}$ range of the CIFAR-100 models. The sign pattern of
$r(\Psi, \mathrm{acc})$ is stable across all weight combinations
$w_H \in \{0.3, 0.5, 0.7\}$ tested (Table~\ref{tab:sens_weights}).

In the language of the dynamical framework, $H_{\mathrm{eff}}$ near unity
indicates that layer activations achieve a Hurst exponent close to
$H_{\mathrm{opt}}$, the empirically optimal regime of persistent
long-range correlations. The CIFAR-100 models remain in a regime where
$H_{\mathrm{raw}}$ deviates substantially from $H_{\mathrm{opt}}$, and
the Gaussian penalty in Eq.~(\ref{eq:heff}) suppresses
$H_{\mathrm{eff}}$ toward zero. This is associated with an apparent
integration ceiling in the configurations studied, representing a maximum
level of cross-layer correlation structure that these models did not
surmount within the training budget. Whether this reflects a genuine
task-imposed constraint or a coincidence of depth, learning rate, and
training duration cannot be determined from single-run observations.

VGG-16 is the instructive exception. It achieves
$\bar{H}_{\mathrm{eff}} = 0.303$, the highest among CIFAR-100 models,
and correspondingly the highest CIFAR-100 accuracy at $63.8\%$. The
intermediate $H_{\mathrm{eff}}$ value is consistent with partial
hierarchical integration, in that the model has found a structured but
incomplete attractor. In dynamical terms this corresponds to a regime of
moderate integration and moderate volatility, sitting between the
high-complexity convergent pattern of DenseNet-121 and the near-zero
integration of the CIFAR-100 ResNets.

The pretrained ViT achieves $\bar{H}_{\mathrm{eff}} = 0.980$, the
highest value across all nine configurations, consistent with the richer
representational prior provided by pre-training on large-scale
data~\cite{dosovitskiy2021image, he2022masked}. This is consistent with
the hypothesis that pre-training raises the apparent integration level
attainable during fine-tuning beyond what scratch training on CIFAR-10
can reach in the studied epoch range.

\subsection{$\Psi$ Volatility as a Convergence Stability Indicator}
\label{sec:psi}

\begin{figure}[t]
  \centering
  \includegraphics[width=\textwidth]{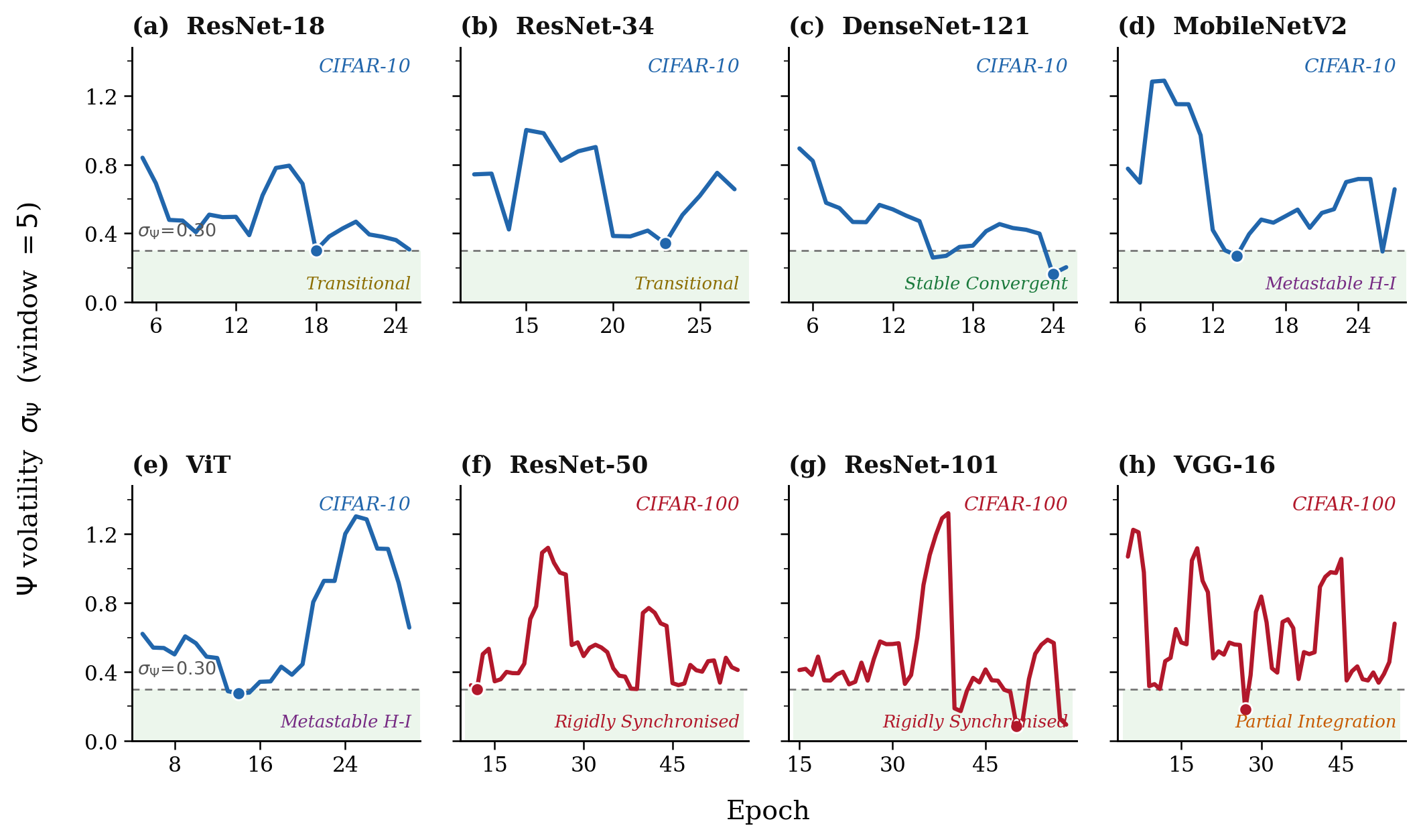}
  \caption{$\Psi$ volatility $\sigma_\Psi$ (rolling standard deviation,
    window $= 5$ epochs) across training for all models. Filled circle
    marks the epoch of minimum volatility. The dashed line denotes the
    candidate convergence threshold $\sigma_\Psi = 0.30$. Green shading
    indicates the convergence zone. CIFAR-10 architectures appear in the
    top row and CIFAR-100 architectures appear in the bottom row. Dataset
    tags and state labels are shown in each panel.}
  \label{fig:psi}
\end{figure}

Figure~\ref{fig:psi} shows the rolling standard deviation of $\Psi$,
denoted $\sigma_\Psi$, for the eight configurations included in the
figure (ResNet-152 is excluded due to insufficient usable $\Psi$ values
for the rolling window, as noted in Table~\ref{tab:results}) over
training epochs. The figure reveals a consistent pattern in which
$\sigma_\Psi$ is elevated in early training and, for several
configurations, collapses toward a lower plateau. We identify a candidate
convergence threshold of $\sigma_\Psi = 0.30$; the sensitivity of
crossing epochs to this choice is reported in
Table~\ref{tab:sens_thresh} (Section~\ref{sec:sensitivity}).

DenseNet-121 provides the clearest example. $\sigma_\Psi$ falls from
$0.89$ at the onset of the rolling window (epoch 5) to $0.20$ at epoch
24, crossing the threshold at epoch 15. This collapse precedes the
accuracy plateau (reached at approximately epoch 22) by seven epochs,
suggesting that $\sigma_\Psi$ may provide an advance signal of
convergence not visible in the loss or accuracy curves. Whether this
anticipation is a reliable property or a coincidence of this single run
cannot be determined from the present data.

ResNet-18 shows a monotone decline from $0.84$ to $0.31$, approaching
but barely crossing the threshold. ResNet-34 oscillates throughout with a
minimum of $0.34$, never fully stabilising. The pretrained ViT shows an
unusual trajectory in which $\sigma_\Psi$ dips below $0.30$ transiently
in epochs 13--15 and then surges to $1.30$ in epochs 24--26 before
settling, reflecting the instability introduced by fine-tuning from a
pre-trained initialisation. Among CIFAR-100 models, ResNet-50 shows the
most unstable volatility trajectory. $\sigma_\Psi$ oscillates between
$0.30$ and $1.12$ throughout training, with no sustained collapse,
indicating the system cannot find a stable attractor. ResNet-101 shows a
distinctive late-stabilisation in which $\sigma_\Psi$ drops to $0.09$ at
epoch 50 before a final surge and recovery, consistent with the model
approaching a low-quality attractor late in training. VGG-16 shows slow
but meaningful decline from $1.07$ to $0.68$ over 55 epochs, suggesting
gradual stabilisation without complete convergence within the training
budget.

These patterns are consistent with the hypothesis that $\sigma_\Psi$
reflects training stability. The sensitivity of threshold-crossing epochs
to the choice of threshold (0.25, 0.30, or 0.35) is reported in
Table~\ref{tab:sens_thresh}; the threshold is configuration-dependent and
no single universal value is supported by this pilot study. A prospective
evaluation with multiple seeds, a pre-fixed threshold, and comparison
against patience-based stopping is required before this criterion can be
recommended.

\subsection{Inter-Field Synchrony and Representational Coherence}
\label{sec:synchrony}

The Pearson correlation between $H_z$ and $M_z$ across epochs,
$r(H_z, M_z)$, partitions the models into two groups. ResNet-50 and
ResNet-101 on CIFAR-100 exhibit strong positive synchrony ($r = 0.864$
and $0.885$ respectively), meaning that integration and metastability
fluctuate together throughout training. ResNet-18 also shows elevated
synchrony ($r = 0.777$). By contrast, DenseNet-121 exhibits negative
synchrony ($r = -0.371$) and the ViT is near-zero ($r = 0.036$).

This separation is theoretically significant. In the source dynamical
framework, states with high dynamical complexity are characterised by
pairwise correlations between components below $0.25$, reflecting
complementary but independent contributions~\cite{ugail2025consciousness}.
When $H_z$ and $M_z$ are tightly locked in phase, increasing integration
comes at the cost of metastability, and vice versa. The system is in a
rigid, low-complexity attractor where the two dimensions cannot vary
independently. The source framework labels this signature as
characteristic of reduced-complexity regimes, and in the present context
it is interpreted as indicating a training attractor with limited
representational flexibility.

In the training context, the high synchrony of ResNet-50 and ResNet-101
is consistent with their failure to escape a low-quality attractor. The
model has found a fixed point where any perturbation to integration
immediately propagates to metastability, leaving the system unable to
explore the representation space flexibly. DenseNet-121's negative
synchrony, by contrast, means that integration and metastability evolve
independently, allowing the system to optimise both dimensions and
converge to a richer attractor.

The finding that the three lowest-performing CIFAR-100 models (ResNet-50,
ResNet-101, VGG-16) are among the four with the highest $r(H_z, M_z)$
supports the interpretation that inter-field synchrony is a diagnostic of
limited representational flexibility within the configurations studied.
DenseNet-121, the only model with negative synchrony, is also the only
one to achieve the Stable Convergent pattern described in
Section~\ref{sec:taxonomy}.

\subsection{Hyperparameter Sensitivity Analysis}
\label{sec:sensitivity}

To assess how the main findings depend on the choice of inherited
parameters, this section reports a systematic recomputation of
$H_{\mathrm{eff}}$ and $\Psi$ across a grid of values for the DFA
parameters ($H_{\mathrm{opt}}$, $\sigma_H$) and the composite weights
($w_H$, $w_M$). All recomputations use the stored raw Hurst exponent
values, reapplying Eq.~(\ref{eq:heff}) with varied parameters. The
analysis cannot substitute for multi-seed replication, but it separates
the effect of hyperparameter choices from the effect of training
stochasticity.

Table~\ref{tab:sens_heff} reports mean $\bar{H}_{\mathrm{eff}}$ across
CIFAR-10 and CIFAR-100 configurations over a $4 \times 4$ grid of
$H_{\mathrm{opt}}$ and $\sigma_H$ values.

\begin{table}[t]
\centering
\small
\caption{Mean $H_{\mathrm{eff}}$ for CIFAR-10 (C10, six configurations)
  and CIFAR-100 (C100, three configurations) across hyperparameter
  combinations. Sep.\ indicates whether the gap exceeds $0.30$.
  Reference values ($H_{\mathrm{opt}}=0.7$, $\sigma_H=0.10$) are bold.}
\label{tab:sens_heff}
\begin{tabular}{@{}rrccc@{}}
\toprule
$H_{\mathrm{opt}}$ & $\sigma_H$ & C10 & C100 & Sep. \\
\midrule
0.5 & 0.05 & 0.037 & 0.283 & No \\
0.5 & 0.10 & 0.275 & 0.522 & No \\
0.5 & 0.15 & 0.496 & 0.715 & No \\
0.5 & 0.20 & 0.652 & 0.820 & No \\
\midrule
0.6 & 0.05 & 0.455 & 0.010 & Yes \\
0.6 & 0.10 & 0.663 & 0.169 & Yes \\
0.6 & 0.15 & 0.802 & 0.386 & Yes \\
0.6 & 0.20 & 0.875 & 0.564 & Yes \\
\midrule
\textbf{0.7} & \textbf{0.05} & \textbf{0.463} & \textbf{0.000} & \textbf{Yes} \\
\textbf{0.7} & \textbf{0.10} & \textbf{0.793} & \textbf{0.024} & \textbf{Yes} \\
0.7 & 0.15 & 0.898 & 0.143 & Yes \\
0.7 & 0.20 & 0.941 & 0.309 & Yes \\
\midrule
0.8 & 0.05 & 0.280 & 0.000 & No \\
0.8 & 0.10 & 0.520 & 0.001 & Yes \\
0.8 & 0.15 & 0.704 & 0.036 & Yes \\
0.8 & 0.20 & 0.810 & 0.135 & Yes \\
\bottomrule
\end{tabular}
\end{table}

The separation holds in 11 of 16 combinations, representing 69 per cent
of the grid. The four combinations where it fails all occur at
$H_{\mathrm{opt}} = 0.5$, where the Gaussian tuning function is centred
below the typical $H_{\mathrm{raw}}$ range of CIFAR-100 models
($H_{\mathrm{raw}} \approx 0.32$ to $0.47$), causing them to receive
higher $H_{\mathrm{eff}}$ scores than the CIFAR-10 models. This reversal
is not observed for $H_{\mathrm{opt}} \geq 0.6$.

Table~\ref{tab:sens_weights} reports the Pearson correlation
$r(\Psi, \mathrm{acc})$ for five representative configurations under
three weight settings with $w_H \in \{0.3, 0.5, 0.7\}$.

\begin{table}[t]
\centering
\small
\caption{Pearson correlation $r(\Psi, \mathrm{acc})$ for five
  configurations under varied composite weights. Bold marks the reference
  setting with $w_H = 0.5$. All values are from single training runs.}
\label{tab:sens_weights}
\begin{tabular}{@{}lrrr@{}}
\toprule
Configuration & $w_H{=}0.3$ & $\bm{w_H{=}0.5}$ & $w_H{=}0.7$ \\
\midrule
DenseNet-121 (C10) & $+0.601$ & $\bm{+0.560}$ & $+0.406$ \\
ViT (C10)          & $-0.323$ & $\bm{-0.329}$ & $-0.291$ \\
ResNet-50 (C100)   & $-0.827$ & $\bm{-0.783}$ & $-0.728$ \\
ResNet-101 (C100)  & $-0.742$ & $\bm{-0.723}$ & $-0.697$ \\
VGG-16 (C100)      & $-0.326$ & $\bm{-0.385}$ & $-0.417$ \\
\bottomrule
\end{tabular}
\end{table}

The sign of $r(\Psi, \mathrm{acc})$ is stable across all weight
combinations for all five configurations tested. DenseNet-121 is the only
configuration with a consistently positive correlation, regardless of how
much weight is placed on integration versus metastability.

Table~\ref{tab:sens_thresh} shows the epoch at which $\sigma_\Psi$
first crosses three candidate threshold levels alongside the epoch of the
accuracy plateau.

\begin{table}[t]
\centering
\footnotesize
\caption{Epoch at which $\sigma_\Psi$ first crosses the stated threshold
  and the accuracy plateau epoch. Dashes indicate the threshold was
  never crossed. All values are from single training runs.}
\label{tab:sens_thresh}
\begin{tabular}{@{}llcccr@{}}
\toprule
Config. & Dataset & $<0.25$ & $<0.30$ & $<0.35$ & Acc.\ plateau \\
\midrule
ResNet-18    & C10  & ---  & ---  & 18   & ep.~22 \\
DenseNet-121 & C10  &  24  &  15  & 15   & ep.~22 \\
MobileNetV2  & C10  & ---  &  14  & 13   & ep.~24 \\
ResNet-101   & C100 &  40  &  40  & 20   & ep.~55 \\
VGG-16       & C100 &  27  &  27  &  9   & ep.~52 \\
ViT          & C10  & ---  &  13  & 13   & ep.~4\rlap{$^\dagger$} \\
\midrule
\multicolumn{6}{l}{$^\dagger$~ViT reaches 99\% of its maximum accuracy by
  epoch 4 due to pretrained initialisation.}\\
\end{tabular}
\end{table}

For DenseNet-121 all three thresholds identify epoch 15, which precedes
the accuracy plateau by approximately seven epochs. ResNet-18 never
crosses the 0.30 threshold despite achieving reasonable accuracy,
illustrating that no single threshold value is universally informative.
These results confirm that the threshold choice is
configuration-dependent and that a universal value cannot be derived
from this pilot study.

\section{Training State Taxonomy}
\label{sec:taxonomy}

The three dynamical signatures of $\bar{H}_{\mathrm{eff}}$,
$\sigma_\Psi$ trend, and $r(H_z, M_z)$ together define a four-state
taxonomy that partitions the nine configurations studied into
qualitatively distinct training regimes. Table~\ref{tab:taxonomy} gives
the formal characterisation of each state. \textbf{Important caveat.}
This taxonomy was induced retrospectively from the same nine
configurations used to derive it. The thresholds and qualitative
descriptions align with the observed configurations by construction. The
taxonomy should be treated as a set of testable hypotheses for future
prospective evaluation, not as a validated classification scheme.

\begin{table*}[t]
\centering
\caption{Proposed taxonomy of training dynamical states. States are
  ordered from highest to lowest representational quality.
  $\sigma_\Psi$ denotes the rolling standard deviation of $\Psi$
  (window $= 5$ epochs).}
\label{tab:taxonomy}
\resizebox{\textwidth}{!}{%
\begin{tabular}{@{}
  p{2.8cm} p{2.2cm} p{2.8cm} p{2.4cm} p{4.2cm} p{2.8cm}
@{}}
\toprule
\textbf{State} & $H_{\mathrm{eff}}$ (late) & $\sigma_\Psi$ trend &
  $r(H_z,M_z)$ & Dynamical interpretation & Observed models \\
\midrule
Stable Convergent &
  $>0.85$, stable &
  Rapidly collapsing &
  $<0$ (decoupled) &
  Convergence to high-complexity metastable attractor &
  DenseNet-121 \\[4pt]
Metastable H-I &
  $>0.85$ &
  Persistently elevated &
  $\approx 0$ (weakly coupled) &
  High integration, no stable attractor found &
  ViT, ResNet-152, MobileNetV2 \\[4pt]
Partial Integration &
  $0.15$--$0.50$ &
  Slowly collapsing &
  $\approx 0$ (weakly coupled) &
  Apparent integration ceiling; structured but incomplete &
  VGG-16 \\[4pt]
Rigidly Synchronised &
  $<0.15$ &
  Flat / non-converging &
  $>0.80$ (tightly locked) &
  Trapped in low-complexity fixed-point attractor &
  ResNet-50, ResNet-101 \\
\bottomrule
\end{tabular}}
\end{table*}

\textbf{Stable Convergent} is the dynamically optimal pattern observed.
DenseNet-121 achieves high $H_{\mathrm{eff}}$, negative inter-field
synchrony, and rapid $\sigma_\Psi$ collapse, corresponding to convergence
into a rich metastable attractor where both integration and metastability
are individually optimised. The positive $r(\Psi, \mathrm{acc}) = +0.60$
is the taxonomic signature of this state, with $\Psi$ rising alongside
accuracy, meaning the model becomes dynamically richer as it learns,
which represents the theoretical optimum in the framework.

\textbf{Metastable High-Integration} is characterised by high
$H_{\mathrm{eff}}$ but persistently elevated $\sigma_\Psi$. The model
achieves strong hierarchical integration but does not find a stable
attractor. The pretrained ViT exhibits this state due to fine-tuning
dynamics, in which the rich pretrained representations maintain high
$H_{\mathrm{eff}}$ but the fine-tuning process does not equilibrate,
producing the characteristic $\sigma_\Psi$ surge in later epochs.
ResNet-152 reaches this state for a different reason. Excessive depth
relative to the task creates optimisation instability, evidenced by the
missing $H_{\mathrm{eff}}$ values in early epochs where DFA window
requirements are not met. In dynamical terms this state is characterised
by high integration and persistently elevated volatility, a combination
indicating the network is exploring a wide region of representation space
without settling into a stable attractor.

\textbf{Partial Integration} is observed exclusively in VGG-16 on
CIFAR-100. The model reaches a plateau in $H_{\mathrm{eff}}$ at
approximately $0.30$, an intermediate value that reflects structured but
incomplete hierarchical integration. The lack of skip connections in
VGG-16 limits cross-layer information
flow~\cite{li2018visualizing}, restricting the model's ability to
establish the fractal activation structure associated with high
$H_{\mathrm{eff}}$. This state corresponds to the most capable CIFAR-100
architecture, suggesting that partial integration is sufficient, and
perhaps optimal, for this task given the available architectures.

\textbf{Rigidly Synchronised} is the lowest-complexity regime observed.
ResNet-50 and ResNet-101 on CIFAR-100 exhibit near-zero
$H_{\mathrm{eff}}$, flat $\sigma_\Psi$ trajectories, and strong
inter-field synchrony. The system is locked into a rigid, low-complexity
attractor where the activation dynamics show little independent variation
across the two components. The strong negative $r(\Psi, \mathrm{acc})$
values ($-0.760$ and $-0.725$ respectively) indicate that as accuracy
rises modestly, $\Psi$ falls further, meaning the model converges toward
an increasingly rigid and low-complexity representational state.

\section{Discussion}
\label{sec:discussion}

The results raise several threads that are worth pursuing, even while
acknowledging that the single-seed, CIFAR-limited nature of the study
prevents strong conclusions.

The most striking pattern is how cleanly the integration measure separates
the two datasets, regardless of which architecture is used. The six
CIFAR-10 configurations consistently achieve high integration scores,
while the three CIFAR-100 configurations remain in a much lower regime.
This is a meaningful observation because it holds across architectures
that differ substantially in design, spanning ResNets, a densely connected
network, a lightweight mobile architecture, and a Vision Transformer,
which makes it less likely to be a quirk of any single design choice. The
sensitivity analysis in Section~\ref{sec:sensitivity} confirms that the
separation is robust to most of the parameter combinations tested, with
the exception of one setting that sits outside the typical operating range
of the framework. If this pattern holds up in multi-seed experiments and
on broader benchmarks, it would suggest that monitoring the integration
measure early in training could give a useful read on whether the network
is on course to develop the internal structure the task appears to require.

Among the CIFAR-10 configurations trained from scratch, DenseNet-121 is
the only one that exhibits the full signature of the Stable Convergent
pattern. It shows high integration, rapidly settling volatility, and
components that fluctuate independently of each other. Its dense
connectivity structure, where every layer receives inputs from all
previous layers~\cite{huang2017densely}, is plausibly related to this
outcome, since it supports richer cross-layer information flow than
architectures that connect only adjacent layers. However, the
configurations differ on several dimensions at once, so this remains a
hypothesis worth testing rather than a conclusion.

The composite index raises a different kind of possibility, namely an
early signal of convergence that does not depend on watching the loss
plateau. When the fluctuation of the composite index settles in the
configurations studied, it often does so before the accuracy curve
flattens. In DenseNet-121 the gap is about seven epochs. The theoretical
intuition is straightforward. If SGD convergence requires gradient
variance to diminish~\cite{robbins1951stochastic}, then a measure of
activation-field variability may carry that signal more directly than the
loss itself. Whether this holds reliably across many architectures and
seeds is the key open question, and answering it properly requires a
prospective study with pre-fixed thresholds and head-to-head comparison
against standard stopping rules.

The four-state taxonomy, for all its appeal as a vocabulary, must be
treated with care. It was constructed by looking at the same nine
configurations used to derive it, so the state assignments are not
independent predictions but descriptive labels fitted to observations.
The connections to established concepts such as flat versus sharp
minima~\cite{keskar2017large}, operation near the edge of
chaos~\cite{langton1990computation}, and criticality and information
capacity~\cite{shew2013functional} are interpretively plausible and worth
following up, but they are post-hoc associations, not confirmed
mechanisms. The taxonomy is best understood as a structured set of
hypotheses that future multi-seed and multi-benchmark experiments could
test and either validate or revise.

\subsection{Limitations}
\label{sec:limitations}

This work has several significant limitations.

All nine configurations were trained once, with a single random seed.
Seed-to-seed variability in $H_{\mathrm{eff}}$, $M$, and $\Psi$ is
unknown. It is possible that the observed state assignments change under
different initialisations. Multi-seed replication, with mean and standard
deviation reported for all dynamical metrics, is the most important
missing element and is the primary direction for future work.

The Hurst exponent is estimated from epoch-indexed activation sequences of
length 25--58 points. This is substantially shorter than typical DFA time
series, for which hundreds to thousands of points are preferred for
reliable long-range correlation estimation~\cite{peng1994mosaic,
kantelhardt2002multifractal}. The reported $H_{\mathrm{eff}}$ values
should be interpreted as coarse indicators of the direction of the
activation correlation structure rather than as precise Hurst exponent
estimates.

The parameters $H_{\mathrm{opt}} = 0.7$, $\sigma_H = 0.1$, and equal
weights $w_H = w_M = 0.5$ are inherited from the source
framework~\cite{ugail2025consciousness}, which calibrated them on
biological EEG signals. The sensitivity analysis in
Section~\ref{sec:sensitivity} shows that the CIFAR-10 versus CIFAR-100
$H_{\mathrm{eff}}$ separation holds for $H_{\mathrm{opt}} \in \{0.6,
0.7, 0.8\}$ with $\sigma_H \geq 0.10$ (11 of 16 combinations tested),
and the $r(\Psi, \mathrm{acc})$ sign pattern is stable across all weight
combinations $w_H \in \{0.3, 0.5, 0.7\}$. However, the absolute values
of all reported metrics are parameter-dependent, and a calibration study
on held-out training runs is necessary before the framework can be
recommended for general use.

The four-state taxonomy was induced from the nine configurations used to
derive it. Additionally, all configurations use CIFAR-10 and CIFAR-100.
Generalisation to larger benchmarks (e.g., ImageNet~\cite{deng2009imagenet}),
other modalities, or generative architectures is untested.

Reducing each layer's activation to a scalar mean per epoch discards
substantial representational structure. The four-layer design provides
some depth coverage but cannot capture within-layer heterogeneity or
geometry that may be relevant for understanding convergence. Richer
signal definitions, including activation variance, the top singular value
of the Gram matrix, or CKA with a reference
layer~\cite{kornblith2019similarity} are also considered as natural
extensions.

\section{Conclusion}
\label{sec:conclusion}

Most studies of deep network training focus on external outcomes such as
loss and accuracy. In this work, we looked at training from a different
angle by asking how the model's internal representations evolve over time
and across layers. By adapting a dynamical complexity framework from
biological signal analysis and applying it to nine model--dataset
configurations, we identified several patterns that are not visible from
standard performance curves alone.

First, the effective integration measure consistently separated
CIFAR-10 from CIFAR-100 across the configurations we studied, and this
pattern remained stable under most of the hyperparameter settings we
tested. Second, the rolling volatility of the composite stability index
often became quiet before accuracy fully levelled off, suggesting that it
may provide an early sign of convergence. Third, the relationship between
integration and metastability appeared to distinguish models that settled
into richer and more flexible training dynamics from those that became
trapped in more rigid and limited regimes.

These observations led us to propose a simple four-state taxonomy of
training behaviour: Stable Convergent, Metastable High-Integration,
Partial Integration, and Rigidly Synchronised. This taxonomy should be
viewed as a descriptive framework rather than a final classification
scheme, since it was derived from the same small set of experiments used
to illustrate it.


\end{document}